\documentclass[11pt,a4paper]{article}
\usepackage[hyperref]{emnlp2020}
\usepackage{graphicx}
\usepackage{times}
\usepackage{latexsym}
\usepackage{array}
\newcolumntype{L}[1]{>{\raggedright\let\newline\\\arraybackslash\hspace{0pt}}m{#1}}
\newcolumntype{C}[1]{>{\centering\let\newline\\\arraybackslash\hspace{0pt}}m{#1}}
\newcolumntype{R}[1]{>{\raggedleft\let\newline\\\arraybackslash\hspace{0pt}}m{#1}}
\usepackage{multirow}

\usepackage{microtype}

\aclfinalcopy 

\title{Towards Ethics by Design in Online Abusive Content Detection}

\author{Svetlana Kiritchenko \qquad \qquad \qquad \qquad Isar Nejadgholi\\
\\
National Research Council Canada \\
\texttt \footnotesize \{svetlana.kiritchenko,isar.nejadgholi\}@nrc-cnrc.gc.ca}

\date{}

\begin{document}
\maketitle
\begin{abstract}

To support safety and inclusion in online communications, significant efforts in NLP research have been put towards addressing the problem of abusive content detection, commonly defined as a supervised classification task. 
 The research effort has spread out across several closely related sub-areas, such as detection of hate speech, toxicity, cyberbullying, etc. 
 There is a pressing need to consolidate the field under a common framework for task formulation, dataset design and performance evaluation.
 Further, despite current technologies achieving high classification accuracies, several ethical issues have been revealed. 
  We bring ethical issues to forefront and propose a unified framework as a two-step process. 
 First, online content is categorized around personal and identity-related \textit{subject matter}s. 
 Second, \textit{severity of abuse} is identified through comparative annotation within each category. 
 The novel framework is guided by the Ethics by Design principle and is a step towards building more accurate and trusted models.   
\end{abstract}

\section{Introduction}

With the increased use of social media, especially among young people, serious concerns about safety and inclusion in online communications have been raised. 
Up to 40\% of users have reported being subjected to online harassment, cyberbullying, and other types of abuse \cite{harassment2017,cyberbullying2020}. 
Often, the victims of online abuse are the most vulnerable parts of society: ethnic minorities, LGBTQ community, or people with disabilities. 
In response, many social media platforms strive to monitor online content and quickly remove abusive posts, but the sheer volume of posts poses significant problems. 
Automatic detection of abusive content can provide assistance and (partially) alleviate the burden of manual inspection.

Much NLP research has been devoted to the problem of automatic abusive content detection.\footnote{
We use the term \textit{abusive} broadly, covering the full range of inappropriate and disturbing content, from simple profanities and obscene expressions to threats and severe insults.} 
It has been studied under a plethora of names, such as detection of flaming \cite{spertus1997smokey}, cyberbullying \cite{dadvar2013improving}, online harassment \cite{golbeck2017large}, hate speech \cite{djuric2015hate,davidson2017automated}, toxicity \cite{dixon2018measuring,aroyo2019crowdsourcing}, and others. 
While these 
sub-areas of the general space of abusive language tackle similar problems, they differ in their focus and scope. 
Recent surveys by \citet{schmidt2017survey,fortuna2018survey,mishra2019tackling,vidgen-etal-2019-challenges,vidgen2020directions,Salawu2020} summarize the advancements in these areas focusing mostly on the technical issues and the variety of machine learning approaches proposed for the tasks. 

In this paper, we examine the general area of abusive language detection from the ethical viewpoint.
We bring together all the related sub-fields, and survey the past work focusing on the different formulations of the task and the common data collection and annotation techniques. 
We discuss challenges that the field faces from the ethical perspective, including fairness and mitigation of unintended biases, transparency and explainability, safety and security. 
In accordance with the Ethics by Design principle, we propose to bring the ethical issues to the early stages of system development: the task formulation and data collection and annotation. 

Typically, the task has been defined as a binary classification problem (abusive vs. non-abusive). However, the social and ethical implications of the task call for more fine-grained labelling of abuse. 
A few attempts have been made to separate abusive language into sub-categories, such as hate speech, threats, aggressive, or offensive language, but the obscure boundaries between the sub-categories make the task challenging even for human annotators \cite{poletto2017hate,founta2018large}. 
To overcome this challenge, we propose to annotate abusive texts for severity of abuse using comparative annotation techniques. 
We argue that severity is a simpler yet more practical dimension for abuse categorization. 
Further, employing comparative annotation would generate fine-grained (or continuous) severity scores while improving the overall reliability of annotations \cite{goffin2011all,KiritchenkoM2017bwsvsrs}. 

Besides severity of abuse, the subject matter (or the target of abuse) is a crucial aspect of an abusive text \cite{waseem-etal-2017-understanding,vidgen-etal-2019-challenges,zampieri-etal-2019-predicting}.  
Knowing if the text (abusive or non-abusive) talks about an individual, a group of people, or an entity closely associated with a specific identity group (e.g., Muslims, gay people, etc.) can help in measuring and mitigating unintended biases in data and model outputs, and in transparency and explainability of the models. 

Overall, 
the contributions of this work are as follows:
\begin{itemize}
    \item We survey the existing works on abusive language detection (including various sub-areas, such as detection of hate speech, cyberbullying, online harassment, etc.) focusing on task formulation and data collection and annotation methods;
    \vspace{-2mm}
    \item We enumerate the main challenges of the task with regard to ethical issues, such as fairness, explainability, and safety;
    \vspace{-2mm}
    \item We propose a novel framework comprising two dimensions, severity of abuse and subject matter of an utterance, and outline the ways in which data collection and annotation can be adjusted to the framework. While identifying target groups subjected to abuse have been explored in previous work, annotating for severity of abuse using comparative annotation techniques has not been considered before. We further discuss how the proposed framework can help in addressing the technical and ethical issues. 
\end{itemize}
\vspace{-1mm}
\noindent We focus on abuse detection in online texts, though the framework can be applied to other media (images, video, speech) and multimodal contexts.

\section{Overview of the Common Practices}

We start with reviewing the common practices of formulating the task of online abuse detection and the methods for collecting and annotating datasets.

\subsection{Task Formulation for Automated Abuse Detection}

The abusive content detection task has typically been defined as a supervised classification problem across various definitions and aspects of abusive language.  
In addition to the main task of determining whether a text is abusive or not, several other dimensions have been explored, including expression of abuse, target of abuse, and legality of abuse \cite{waseem-etal-2017-understanding,fivser2017legal,poletto2017hate,vidgen-etal-2019-challenges,niemann2019abusive,zufall2020operationalizing}.

\noindent \textbf{Expression of abuse:} Online abuse can be expressed in different forms, such as hate speech, insults, physical threats, stereotyping, and more. 
Focusing on slightly different aspects of abuse, these categories have obscure boundaries, and are often challenging for humans and machines to split apart \cite{poletto2017hate,founta2018large}.  

Another practical distinction is whether the abusive language is explicit or implicit \cite{waseem-etal-2017-understanding}. 
Explicit abuse is relatively easy to recognize as it contains explicit obscene expressions and slur. 
Implicit abuse, on the other hand, is not immediately apparent as it can be obscured by the use of sarcasm, humor, stereotypes, ambiguous words, and lack of explicit profanity.
In many existing datasets implicit abuse can be found in only a small proportion of instances \cite{wiegand2019detection}. 
Furthermore, implicit abuse presents additional challenges to human annotators as sometimes specific background knowledge and experience are required in order to understand the hidden meaning behind implicit statements. 
To deal effectively with this class of abuse, large annotated datasets focusing on implicitly abusive language are needed so that automatic detection systems are exposed to a wide variety of such examples through their training data.

\noindent \textbf{Target of abuse:} Abusive speech can be directed towards particular person(s) or entities, or contain undirected profanities and indecent language \cite{zampieri-etal-2019-predicting}. 
While obscene language, in general, can be disturbing to some audiences, abuse targeting specific individuals or groups is often perceived as potentially more harmful and more concerning for society at large. 
Therefore, majority of research on abusive language detection has been devoted to targeted abuse. 
\citet{waseem-etal-2017-understanding} distinguished two target types: an individual and a generalized group. 
They argued that the distinction between an attack directed towards an individual or a generalized group is important from both the sociological 
and the linguistic points of view. 
Thus, this distinction may call for different handling of the two types of abusive language when manually annotating abusive speech and when building automatic classification systems. 
For example, in research on cyberbullying, where abusive language is directed towards specific individuals, more consensus in task definition and annotation instructions can be found, and higher inter-annotator agreement rates are often observed \cite{dadvar2013improving}. 

A third target type---entity or concept---can also be considered \cite{zampieri-etal-2019-predicting,vidgen-etal-2019-challenges}. Acceptable criticism of an entity (e.g., country), a concept (e.g., religion), an organization, or an event, can be semantically similar to abusive language.
However, there is often a thin line between criticizing a concept and attacking people associated with the concept (e.g., an anti-Islamic propaganda can induce hatred towards Muslims).

\noindent \textbf{Legality of abuse:} Some types of abusive statements, such as hate speech and defamatory allegations, are not only morally unacceptable, but also illegal in several countries. 
To automatically determine if a statement is illegal, the corresponding laws need to be translated into manageable NLP tasks \cite{fivser2017legal,zufall2020operationalizing}. 
However, the definitions of illegal online abuse vary across jurisdictions and typically cover only the most severe cases of abuse that can threaten the society at large. 
Therefore, the NLP research community should focus on a broader problem and design solutions that can be easily configurable for a specific set of requirements.

\subsection{Data Collection}
Several datasets manually annotated for abuse detection have been made available. Datasets can be collected from a single platform, such as Yahoo! \cite{djuric2015hate}, Wikipedia \cite{wulczyn2017ex}, Facebook \cite{kumar2018benchmarking}, Twitter \cite{waseem2016hateful,davidson2017automated,founta2018large}, or from multiple discussion forums \cite{van2020multi}.

Generally, it is laborious and costly to build an abuse detection corpus that is balanced with respect to hateful and harmless comments \cite{schmidt2017survey}. Since abusive behaviour is relatively infrequent, random sampling results in datasets extremely skewed towards benign samples \cite{founta2018large}. Existing sampling strategies rely on known abusive/profane words, words describing the target populations, or monitoring users known for abusive behavior. 
A combination of random sampling and targeted search have also been used \cite{chatzakou2017mean,wulczyn2017ex,founta2018large}.

Sometimes, specific data collection procedures are defined based on the task at hand. 
For example, \citet{hosseinmardi2015analyzing} used a snowball sampling method starting from a small number of users who posted offensive content on Instagram. 
\citet{waseem2016hateful} focused on sexism and racism, and collected tweets matching query words that are likely to occur in these cases. \citet{davidson2017automated} used a lexicon of words and phrases identified by users as related to hate speech. 

\subsection{Data annotation}

\citet{tversky1974judgment} were the first psychologists that showed how humans employ heuristics to make judgements under uncertainty. These heuristics are formed based on complex factors and lead to systematic personal biases. On top of the general issue of subjectivity, in the case of abuse detection, 
a different understanding of what to consider abusive language resulted in sometimes contradictory annotation guidelines, and incompatible and erroneous datasets. For example, \citet{van2018challenges} questioned 10--15\% of manually obtained labels on two widely used datasets, Kaggle Toxicity by Jigsaw and Google\footnote{https://www.kaggle.com/c/jigsaw-toxic-comment-classification-challenge} and the one by \citet{davidson2017automated}. \citet{waseem-etal-2017-understanding} and \citet{nobata2016abusive} observed that expert annotators reach higher inter-rater agreements and produce better quality annotations compared to crowd-sourced workers.

To minimize the effect of subjectivity, some of the datasets are annotated by multiple annotators. 
The proportion of majority votes per instance represents the level of agreement, and can serve as a rough estimate for severity of abuse.  
However, most often the votes are aggregated into a single label. \citet{wiegand2019detection} and \citet{davidson2017automated} used majority voting whereas \citet{gao2017detecting} annotated a statement as hate speech if at least one annotator labeled it as hateful. \citet{golbeck2017large} collected judgements from two trained annotators, and a third annotator was employed only if the first two disagreed.

\section{Current Technical and Ethical Challenges}
In this section, we highlight common technical and ethical issues related to the currently employed task definition and the available datasets, regardless of applied machine learning techniques.

\noindent \textbf{Task Formulation:} 
In practical applications, the definitions of abusive language heavily rely on community norms and context and, therefore, are imprecise, application-dependent, and constantly evolving  \cite{chandrasekharan2018internet}. 
To make the task more tractable and focused, previous research has mostly concentrated on specific types of online abuse (e.g., hate speech, sexism, personal attacks). 
However, it has been shown that defining and annotating types of abuse are challenging tasks and often result in inconsistent definitions across studies, highly overlapping categories, and low inter-annotator agreements \cite{waseem-etal-2017-understanding,poletto2017hate,founta2018large}. 

Recently, the complexity of the task formulation has been brought to the attention of the community, and  several studies have proposed multi-level frameworks to address the task.  
\citet{waseem-etal-2017-understanding} mapped the different types of abuse to two-dimension: identity- versus person-directed abuse and explicit versus implicit abuse. They argued that inter-annotator agreement is high when the abuse is directed to a person and explicit and low when the abuse is generalized and implicit. 
\citet{fortuna-etal-2020-toxic} demonstrated that many different definitions are being used for equivalent concepts, which makes most of the publicly available datasets incompatible. They suggested that hierarchical multi-class annotation schemas should be deployed to formulate the online abuse detection task. \citet{sap2020socialbiasframes} formulated offensive language detection as a hierarchical task that combines structured classification with reasoning on social implications. They trained a model that translates an offensive statement to the implied stereotype that is hurtful to the target demographic. \citet{assimakopoulos-etal-2020-annotating} formulated hate speech as hierarchical and multi-layer inferences on sentiment, target, expression of abuse and violence.  

Because of the complexities of task formulation, most of the studies focus on one specific dataset, and combining existing datasets is not a trivial task.
Moreover, the scope of studied abusive behaviors has been limited \cite{jurgens-etal-2019-just}.

\noindent \textbf{Sampling Bias:} 
Sampling techniques deployed to boost the number of abusive examples may result in a skewed distribution of concepts and entities related to targeted identity groups. These unintended entity misrepresentations often translate into biased abuse detection systems.  
\citet{dixon2018measuring} and \citet{davidson2019racial} focused on the skewed representation of vocabulary related to racial demographics in the abusive part of the dataset, and showed that adding counter-examples (benign sentences with the same vocabulary) would mitigate the bias to some extent. \citet{park2018reducing} measured gender bias in models trained on different abusive language datasets and suggested various mitigation techniques, such as debiasing an embedding model, proper augmentation of training datasets, and fine-tuning with additional data.  
\citet{nejadgholi2020cross} explored multiple types of selection bias and demonstrated that the ratio of offensive versus normal examples leads to a trade-off between False Positive and False Negative error rates. They concluded that this ratio is more important than the size of the training dataset for training effective classifiers. They also showed that the source of the data and the collection method can lead to topic bias and suggested that this bias can be mitigated through topic modeling.  

\noindent \textbf{Annotation Bias:} Besides skewed data representations resulting from data sampling, annotator bias is another barrier for building fair and robust systems. \citet{wilhelm2019gendered} studied the influence of social media users’ personal characteristics on the evaluation of hate comments, focusing on abuse aimed towards women and sexual minorities. Their results indicate that moral judgments can be gendered.  \citet{breitfeller-etal-2019-finding} used the degree of discrepancies in annotations between male and female annotators to surface 
nuanced microaggressions. 
Also, it has been shown that annotators' knowledge of different aspects of hateful behaviour can have a significant impact on the performance of trained classification models \cite{waseem2016you}. Similarly, annotators’ insensitivity or unawareness of dialect can lead to biased annotations and amplify harms against racial minorities \cite{waseem2018bridging,sap2019risk}. 

\noindent \textbf{Quantifying Bias:} Even though the developers of datasets and models are cognizant of the risk of various biases, quantifying the extent of this risk is challenging. \citet{dixon2018measuring} proposed a way of measuring bias in trained models by building a synthetic dataset and using an evaluation metric that computes error disparity across identity groups. Kaggle competition on the Unintended Bias in Toxicity Classification, introduced a set of metrics that measure unintended bias for identity references across multiple dimensions. Also, different definitions and frameworks of fairness have been used for the evaluation of automatic abuse detection systems to encourage the development of systems that are optimized not only for the overall performance but also for fair outputs across different target groups \cite{borkan2019nuanced, garg2019counterfactual}. \citet{dinan2020multi} decomposed gender bias in text along several pragmatic and semantic dimensions and proposed classifiers for controlling gender bias. 

Embedding models are one of the important sources of bias in natural language processing systems. There has been an active line of work that aims to quantify bias and stereotypes in language models that generate text representations. Early works focused on gender and racial bias and introduced association tests for measuring bias in word embeddings \cite{bolukbasi2016man,caliskan2017semantics,manzini2019black}. For contextualized word embeddings, \citet{may2019measuring} and \citet{kurita2019measuring} used pre-defined sentence templates, and \citet{nadeem2020stereoset} and \citet{nangia2020crows} collected crowd-sourced sentences to measure stereotypical biases hidden in language models. Not only do pre-trained neural language models reflect social biases, they are also prone to generating racist, sexist, or otherwise toxic language which hinders their safe deployment \cite{gehman2020realtoxicityprompts}. However, it is still unclear how the bias and toxicity present in language models impact the output of the trained classifiers.

\noindent \textbf{Generalizability:} \citet{wiegand2019detection} showed that sampling bias can limit the generalizability of trained models. 
Depending on the sampling method and the platform that the dataset is collected from, some datasets are mostly comprised of explicitly abusive texts while others mainly contain sub-types of implicit abusive language such as stereotypes. The study demonstrated that models trained on datasets with explicit abuse and less biased sampling perform well on other datasets with similar characteristics, whereas datasets with implicit abuse and biased sampling contain specific features usually not generalizable to other datasets. \citet{nejadgholi2020cross} demonstrated that platform-specific topics can negatively impact the generalizability of the trained classifiers. They showed that removing over-represented benign topics can improve the generalization across datasets. 

\noindent \textbf{Explainability:} 
As the impact of AI becomes more significant in our daily lives, developers of automatic systems are expected to earn the trust of users by providing explanations for automatically made decisions. Traditional lexicon and feature-based models are interpretable to some extent as they use features understandable by humans. 
Several lexicons of abusive expressions have been built manually, automatically, and semi-automatically \cite{razavi2010offensive,gitari2015lexicon,wiegand2018inducing}. 
In feature-based systems, bag-of-words and character n-grams have been most frequently used, but some other explainable features, such as the ones derived from sentiment analysis, tone analysis, subjectivity, and topic modelling, have also been employed 
\cite{fortuna2018survey}. However, the accuracy of lexicon and feature-based systems is often significantly lower than the accuracy of deep learning models \cite{dixon2018measuring,gunasekara2018review,founta2019unified}.

Neural networks, on the other hand, are effectively black boxes. 
Recent research has leveraged the LIME (locally interpretable model-agnostic explanations) algorithm in an attempt to interpret a model's representation of abusive statements \cite{srivastava2019detecting,mahajan2020explainable}. 
LIME's explanations consist of words highly weighted by the model, but no further information is provided on why a text is classified as abusive \cite{ribeiro2016should}. 
Similarly, attention mechanisms embedded in deep learning architectures were used to identify the abusive parts of a text \cite{chakrabarty-etal-2019-pay}. 
However, it is not clear if such mechanisms provide meaningful explanations of a model's decisions \cite{jain-wallace-2019-attention,wiegreffe2019attention}.

Output probability (or confidence) scores produced by classifiers have been used to explain the severity of abuse \cite{hosseini2017deceiving,grondahl2018all}. However, it is not clear how well these probabilities correspond to the human perception of severity and in what ways they might be affected by sampling methods.

Another approach to explainability is through more comprehensive data annotation so that more particulars 
can be learned directly from training data. For example, models trained on the Kaggle Toxicity dataset labelled for five sub-categories of toxicity provide more information than the previous versions of this dataset annotated with binary labels 
\cite{wulczyn2017ex}. Another example is the OffensEval dataset that includes annotations for the target of abuse (individual, group, or other) 
\cite{zampieri2019semeval}. \citet{sap2020socialbiasframes} employed modern large-scale language models in an attempt to automatically generate explanations as social bias inferences for abusive social media posts that target members of identity groups. 
They asked human annotators to provide free-text statements that describe the targeted identity group and the implied meaning of the post in the form of simple  patterns (e.g., ``women are ADJ'', ``gay men VBP'').
They showed that while the current models are capable of accurately predicting whether the online post is offensive or not, they struggle to effectively reproduce human-written statements for implied meaning.

\noindent \textbf{Transparency:} \citet{mitchell2019model} introduced the concept of model cards as a means to address transparency of deep learning models. 
They urged the developers of models to report the details of data on which the models were trained and clarify the scope of use, including the applications where the employment of the model is not recommended. 
As an example, they presented a model card for an automatic abuse detection system, Perspective API \cite{PersAPI}. 
Similar concepts, data statements \cite{bender2018data} and datasheets for datasets \cite{Gebru2018DatasheetsFD}, were proposed to standardize the process of documenting datasets.
\citet{bender2019typology} explained how transparent documentation can help in mitigating the ethical risks.

\noindent \textbf{Safety and Security:} Several studies have shown that trained abuse detection systems can be deceived or attacked by malicious users. \citet{hosseini2017deceiving} demonstrated that an adversary can query the system multiple times and find a way to subtly modify an abusive phrase resulting in significantly lowering confidence that the phrase is abusive. 
\citet{grondahl2018all} showed that adding a positive word such as \textit{`love'} to an abusive comment can flip the model’s predictions. They studied seven models trained for hate speech detection and concluded that although character-based models are more resistant to attacks, model variety is less important than the type of training data and labelling criteria. 
Further, \citet{kurita2020weight} observed that in spite of rich sub-word representations, a BERT-based classifier can be deceived by inserting a specific rare word to an abusive sentence. \citet{kalin2020systematic} proposed a structured approach for securing a toxicity detection classifier in a production setting.

\section{A Novel Framework for Abusive Language Detection}

To address some of the issues outlined in the previous section, we propose a novel framework for categorizing online abusive language. 
We identify two primary dimensions of interest, designed to cover the full range of the spectrum, from most genial, non-abusive to extremely abusive texts: 
 
\vspace{-2mm}
\begin{enumerate}

    \item subject matter of utterance
    \vspace{-2mm}
    \item severity of abuse
\end{enumerate}
\vspace{-2mm}

\noindent \textbf{Subject Matter:} As previous research demonstrated, it is essential from both legal and linguistic points of view to identify the target of abuse---who (or what) the abuse is directed towards. 
We extend this idea to cover both abusive and non-abusive texts, and propose to specify the subject matter of an utterance. 
We define subject matter as the topic of a factual statement or the target of an opinion. 
Having all types of texts, positive, negative, and neutral opinions as well as factual statements, annotated for subject matter can help in balancing the distribution of abusive and non-abusive instances for different types of targets and broadening the range of instances related to the same subject matter. 
However, we consider only subject matters that can potentially be targets of abuse, namely people (individuals and groups) and entities related to identity groups. 
All other subject matters are grouped under the category `other'. 
Identity groups are sections of population with significant membership that are usually defined by ethnicity, religion, gender, or sexual orientation, but can also be defined by other characteristics, such as physical appearance, occupation, political affiliation, etc. 
Entities related to identity groups include concepts (e.g., Islam is related to Muslims), events (e.g., Pride Parade is related to the LGBTQ community), ideas, etc. 
While negative remarks towards entities would normally constitute an acceptable form of criticism, having such instances in the training datasets would ensure the systems' exposure to examples of non-abusive texts linguistically similar to abusive instances.

\begin{figure}[t]
\begin{center}

\includegraphics[scale=0.5]{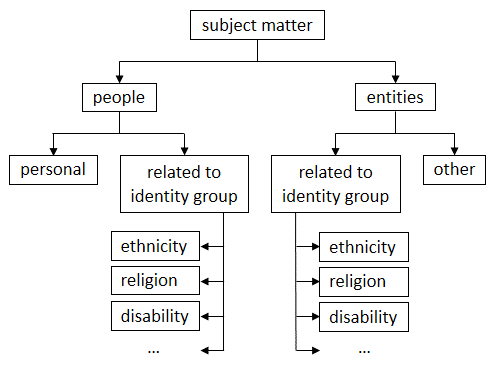} 
\caption{Taxonomy for subject matter of an utterance.}
\label{fig-target}
\end{center}
\end{figure}

When the subject matter is people, we distinguish personal and identity group related reference. 
If the subject matter (whether a single person or a group) is referred to by identity terms associated with an identity group, we call it `related to identity group'; otherwise, we classify it as `personal'.   
Notice that unlike previous research, 
we do not distinguish subject matters at the level of individuals and groups as similar legal considerations and linguistic patterns are observed for both types.  
Identity group related subject matters can be further characterized by the basis on which the identity group is defined (e.g., race)
as well as by the specific identity (e.g., African-Americans).

Figure~\ref{fig-target} shows the full multi-level hierarchical taxonomy for subject matters. 
Since an utterance can refer to more than one identity group, 
multiple categories can be assigned to an instance.

\noindent \textbf{Severity of Abuse:} 
Online abusive content embodies a spectrum of practices that differ in motivation, expression, and consequences \cite{shepherd2015histories,pohjonen2017extreme}. 
While separating different forms of online abuse (e.g., threats, insults, hate speech) proved problematic, a more attainable, yet valuable objective  can be determining the level of severity of abuse---a point on the scale from non-abusive, friendly instances to extremely abusive, violent messages. 
Ordering textual instances by the severity of abuse can help human moderators to prioritize messages for manual inspection and to promptly respond to potentially dangerous ones.

The common technique for annotating items on a fine-grained ordinal scale is rating scales. 
However, traditional rating scales suffer from a number of shortcomings,  including inconsistencies in annotations by different annotators and by the same annotator over time, scale region bias, and fixed granularity  \cite{baumgartner2001response,presser2004questions}. 
To overcome these problems, human annotators can be asked to provide comparative judgements instead \cite{goffin2011all,aroyo2019crowdsourcing}.

An efficient comparative technique, widely used in marketing research, is Best--Worst Scaling \cite{Louviere_1990,Louviere2015}.  
In Best--Worst Scaling (BWS), an annotator is presented with $n$ items (
where $n$ is typically 4 or 5) and asked to select the \textit{best} item (\textit{the most abusive}) and the \textit{worst} item (\textit{the least abusive}). 
All the items to be annotated are organized in $m$ $n$-tuples in such a way that ensures each item is annotated multiple times and compared with a diverse set of other items. After annotating around $1.5 \times N$ to $2 \times N$ such $n$-tuples (where $N$ is the total number of textual instances to be annotated), a real-valued score of severity can be calculated for each textual instance,
and a ranked list of instances by severity can be obtained \cite{flynn2014,maxdiff-naacl2016}. 

Since BWS tuples are formed randomly (though, ensuring the diversity of comparisons for each item), for some tuples the choice of the most abusive and/or least abusive texts might be apparent for most annotators. 
Extremely abusive texts would be often selected as the \textit{most abusive} and get a high severity score while genial and friendly texts would mostly be selected as the \textit{least abusive} and get low scores. 
Yet, in some tuples two or more texts might express similar levels of severity. 
In such cases, since annotators are forced to make a decision for each tuple, the answer by each annotator would be selected randomly between these similarly abusive items. 
This means that the items would be chosen (on average) by the same number of annotators, and, therefore, the aggregated scores for these items would be close to each other. 

\begin{figure}[t]
\centering
\includegraphics[width=1\linewidth]{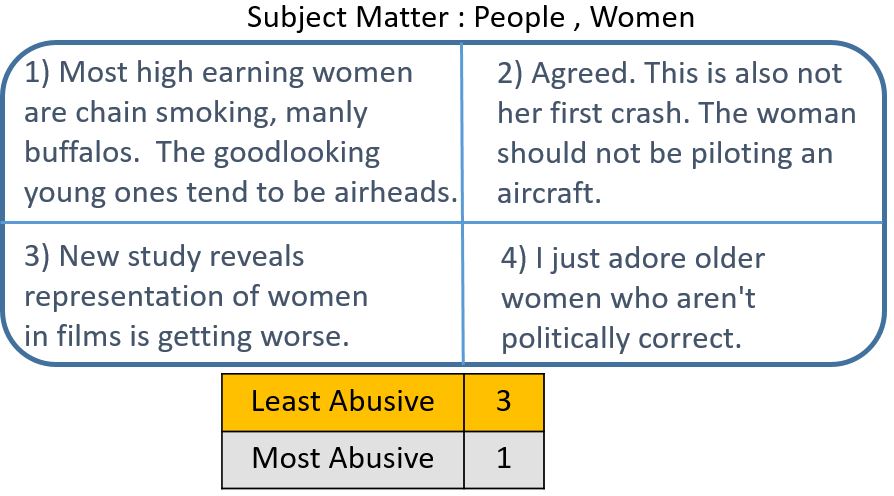}
\caption{Example of a BWS annotation for severity. }
\label{fig:annot}
\end{figure}

It has been shown that Best--Worst Scaling produces more reliable annotations as compared to the traditional rating scales, especially 
on linguistically more complex items \cite{KiritchenkoM2017bwsvsrs}. Figure~\ref{fig:annot} demonstrates an example of a hypothetical BWS annotation for severity of abuse. 
Table~\ref{tab:examples} shows examples of texts annotated according to the entire framework.

\setlength{\tabcolsep}{7pt}
\begin{table*}
\centering
\begin{tabular}{L{10cm}L{2.5cm}c}

{\small \textbf{Text}} &  {\small \textbf{Subject matter }}&  {\small \textbf{Severity score}}  \\
\hline
{\small Claiming to be transgender is a sign of mental illness. The military has no place for people with mental disabilities.} &  {\small People, Transgender} & {\small 0.8} \\
\hline
{\small Transgender people have a gender identity or gender expression that differs from their sex assigned at birth.}  & {\small People, Transgender} & {\small 0.0} \\
\hline

{\small Shove it up your f*cking *ss and burn in hell.} &  {\small People, Personal}&{\small 0.9} \\

\hline
{\small This movie was a f*cking piece of sh*t.} & {\small Entities, Other}& {\small 0.5} \\

\hline

{\small I personally believe that Islam requires a Reformation or an Enlightenment.} & {\small Entities, Islam }& {\small 0.2} \\

\hline

\hline
\end{tabular}
\caption{Examples of texts with hypothetical annotations. Severity scores are on the scale from 0 (least abusive) to 1 (most abusive). The absolute values of severity scores are not meaningful but the relative values distinguish between facts, criticism, obscene sentences, and highly abusive contents. Some of the texts are taken from the Kaggle Unintended Bias in Toxicity dataset by Jigsaw and Google.}
\label{tab:examples}
\end{table*}

\setlength{\tabcolsep}{6pt}

\section{Advantages of the Novel Framework}
\label{sec-advantages}

The proposed framework addresses some of the problems outlined earlier. 

\noindent \textbf{Task Formulation:} The framework focuses on a general class of abusive behavior and aims to extend the scope of studied online abuse without the painful process of enumerating and precisely defining a myriad of types. 
Instead, many types can be roughly defined as regions in the proposed two-dimensional space of subject matter and severity of abuse. 
For example, `physical threat' can be mapped to (subject matter: people, severity: high), and `personal attack' can be mapped to (subject matter: personal, severity: moderate to high). 
In this way, separate research efforts are expected to produce more compatible and more applicable outputs. 
Still, listing different types of abuse with their coarse-grained definitions and examples can be extremely helpful in guiding data collection and annotation.

\noindent \textbf{Data Collection and Sampling Bias:} 
To ensure adequate representation for different identity groups, data can be collected for each group using abusive and benign query terms that refer to members of that group or to entities associated with the group. 
The sets of terms can be acquired manually or semi-automatically using unsupervised techniques, such as topic modeling, clustering, etc. 
Textual messages collected in this way would include abusive as well as neutral and friendly utterances about an identity group or closely related concepts, cover a variety of topics, and contain explicit and implicit language. 
However, this method of data collection does not specifically target the `personal' category. 
It can also result in low proportions of abusive texts and introduce unintended biases, such as topic bias \cite{wiegand2019detection}. 
Therefore, the data collection should be spread out over a period of time (to diversify the set of covered topics) and supplemented with other techniques, such as sampling based on lexicons of common abusive words and expressions, sampling of messages written by users known for abusive behavior, and random sampling. 
Further, data from existing abusive language datasets as well as abusive examples reported by users on dedicated websites, such as HeartMob\footnote{https://iheartmob.org} and Microaggressions\footnote{https://www.microaggressions.com}, are valuable  data sources.  
Collecting data with a specific focus on identity groups allows to account for fairness in representation at the beginning of the development cycle.

The set of identity groups represented in a dataset is decided apriori based on a research focus, a data source, and available resources. 
A preliminary round of exploratory annotations can be beneficial to expand the list of commonly addressed groups (e.g., women, African-Americans,
immigrants) to other identities.   
Further exploratory rounds can be run periodically to include new, previously non-existent or missed, categories. 

\noindent \textbf{Annotation Bias:} 
Manually annotating for severity of abuse is a particularly subjective task, and often requires specific expertise obtained through training or life experiences. 
We recommend having data annotated for subject matter first, and then the severity annotations can be done independently for different identity groups and involve corresponding annotator pools. 
Most often, online abuse is directed towards minorities and marginalized communities, therefore it is vital to involve and consult the members of 
these communities to adequately represent their values and to reduce data annotation bias \cite{blackwell2017classification}. 
In cases where community involvement in annotation is infeasible, professionals specializing in related issues or trained annotators can be employed.

\noindent \textbf{Quantifying Bias:} Although the new framework does not guarantee the fairness of trained models, it allows  
measuring and mitigating bias through comparability of the overall automatic detection error or the False Positive and False Negative error rates for different identity groups.

\noindent \textbf{Generalizability:} 
The conceptions of target groups based on specific identities are expected to be applicable across online platforms and domains. 
The generalizability of trained models will be improved by increasing the proportion of texts with implicit meaning and texts that provide acceptable criticism of entities and concepts. 
Further, the proposed framework will improve the generalizability by allowing data annotators to use the full spectrum of the severity dimension without forcing them to decide where the boundary between abusive and benign languages lies. That decision is application and domain specific and can be left to human moderators, making trained systems suitable to a greater variety of applications.

\noindent \textbf{Explainability:} Within the framework, the systems are trained to predict and output the detailed information on the target of abuse and the level of severity that can serve as basic explanations for human and machine decisions. 
More comprehensive explanations can sometimes be derived when the targeted group is coupled with the words and expressions in the message that are considered particularly insulting for that group. 
Similarly, knowing the target of abuse is imperative in recognizing and explaining implicit abuse and stereotypical references. 
Further, these explanations can serve educational purposes when employed in a system that assists users at the time of message creation.

\noindent \textbf{Transparency:}
The framework provides a means to measure the class imbalances in a dataset across target groups. 
Reporting the limitations of a dataset is essential for transparency and helps practitioners to better identify the scope of use of trained models.

\noindent \textbf{Safety and Security:} 
The vulnerability of classification systems due to their high sensitivity to specific words suggests that those words are over-represented in the training datasets. The proposed framework can improve the safety of such systems by including more training examples with implicit meaning and training systems to learn fine-grained information which is not directly correlated with explicitly abusive words.

\section{Limitations and Ethical Considerations}

Delineating target categories based on identity groups, selecting search terms associated with the groups, and annotating for fine-grained target categories can propagate harmful stereotypes and reinforce social iniquity \cite{beukeboom2019stereotypes}. 
NLP researchers should ground their work in the relevant literature from other disciplines, such as sociology, sociolinguistics, and social psychology, and engage with the lived experiences of members of affected communities in order to minimize such adverse effects \cite{blodgett-etal-2020-language}.

Viewing and annotating abusive content for prolonged periods of time can cause significant distress to human annotators \cite{roberts2016commercial,vidgen-etal-2019-challenges}.  
This concern is even more critical for members of marginalized groups annotating abusive texts directed towards their community. 
To reduce the possible harmful effects on mental health of the annotators, special procedures can be put in place, including the right consent process, comprehensive instructions, limited exposure time, fair compensation, and mental health support. Annotators should be made aware of why they are labeling such contents and how their work contributes to the safety of online platforms for their communities.

\section{Conclusion and Future Directions}

A new framework structured around the dimensions of \textit{subject matter} and \textit{severity of abuse} is proposed as a step towards building less biased, more accurate and generally more trustable automatic abuse detection systems. 
Comprehensive data collection and annotation proposed within the framework allow for better control and transparency on data characteristics and model performance with regard to unintended biases, generalizability, explainability, and safety. 
As the next step, the usability and efficacy of the proposed framework need to be tested. 
Thus, future work will include building an extensive list of identity groups subjected to online abuse, assembling lexicons of terms associated with the groups for targeted sampling, and writing detailed annotation guidelines for both annotation steps. 
Then, empirical datasets in multiple languages can be collected, annotated, and released to the research community for experimenting and building trustable machine learning models.

\bibliographystyle{acl_natbib}
\bibliography{abusive-lang}

\begin{thebibliography}{92}
\expandafter\ifx\csname natexlab\endcsname\relax\def\natexlab#1{#1}\fi

\bibitem[{van Aken et~al.(2018)van Aken, Risch, Krestel, and
  L{\"o}ser}]{van2018challenges}
Betty van Aken, Julian Risch, Ralf Krestel, and Alexander L{\"o}ser. 2018.
\newblock Challenges for toxic comment classification: An in-depth error
  analysis.
\newblock In \emph{Proceedings of the 2nd Workshop on Abusive Language Online}.

\bibitem[{Aroyo et~al.(2019)Aroyo, Dixon, Thain, Redfield, and
  Rosen}]{aroyo2019crowdsourcing}
Lora Aroyo, Lucas Dixon, Nithum Thain, Olivia Redfield, and Rachel Rosen. 2019.
\newblock Crowdsourcing subjective tasks: the case study of understanding
  toxicity in online discussions.
\newblock In \emph{Proceedings of the Workshop on Subjectivity, Ambiguity and
  Disagreement in Crowdsourcing}, pages 1100--1105.

\bibitem[{Assimakopoulos et~al.(2020)Assimakopoulos, Vella~Muskat, van~der
  Plas, and Gatt}]{assimakopoulos-etal-2020-annotating}
Stavros Assimakopoulos, Rebecca Vella~Muskat, Lonneke van~der Plas, and Albert
  Gatt. 2020.
\newblock \href {https://www.aclweb.org/anthology/2020.lrec-1.626} {Annotating
  for hate speech: The {M}a{N}e{C}o corpus and some input from critical
  discourse analysis}.
\newblock In \emph{Proceedings of the 12th Language Resources and Evaluation
  Conference}, pages 5088--5097, Marseille, France. European Language Resources
  Association.

\bibitem[{Baumgartner and Steenkamp(2001)}]{baumgartner2001response}
Hans Baumgartner and Jan-Benedict~E.M. Steenkamp. 2001.
\newblock Response styles in marketing research: A cross-national
  investigation.
\newblock \emph{Journal of Marketing Research}, 38(2):143--156.

\bibitem[{Bender(2019)}]{bender2019typology}
Emily~M Bender. 2019.
\newblock A typology of ethical risks in language technology with an eye
  towards where transparent documentation can help.
\newblock In \emph{Future of Artificial Intelligence: Language, Ethics,
  Technology Workshop}.

\bibitem[{Bender and Friedman(2018)}]{bender2018data}
Emily~M Bender and Batya Friedman. 2018.
\newblock Data statements for natural language processing: Toward mitigating
  system bias and enabling better science.
\newblock \emph{Transactions of the Association for Computational Linguistics},
  6:587--604.

\bibitem[{Beukeboom and Burgers(2019)}]{beukeboom2019stereotypes}
Camiel~J. Beukeboom and Christian Burgers. 2019.
\newblock How stereotypes are shared through language: a review and
  introduction of the social categories and stereotypes communication ({SCSC})
  framework.
\newblock \emph{Review of Communication Research}, 7:1--37.

\bibitem[{Blackwell et~al.(2017)Blackwell, Dimond, Schoenebeck, and
  Lampe}]{blackwell2017classification}
Lindsay Blackwell, Jill Dimond, Sarita Schoenebeck, and Cliff Lampe. 2017.
\newblock Classification and its consequences for online harassment: {D}esign
  insights from {H}eart{M}ob.
\newblock \emph{Proceedings of the ACM on Human-Computer Interaction},
  1(2):1--19.

\bibitem[{Blodgett et~al.(2020)Blodgett, Barocas, Daum{\'e}~III, and
  Wallach}]{blodgett-etal-2020-language}
Su~Lin Blodgett, Solon Barocas, Hal Daum{\'e}~III, and Hanna Wallach. 2020.
\newblock Language (technology) is power: A critical survey of {``}bias{''} in
  {NLP}.
\newblock In \emph{Proceedings of the 58th Annual Meeting of the Association
  for Computational Linguistics}, pages 5454--5476.

\bibitem[{Bolukbasi et~al.(2016)Bolukbasi, Chang, Zou, Saligrama, and
  Kalai}]{bolukbasi2016man}
Tolga Bolukbasi, Kai-Wei Chang, James~Y Zou, Venkatesh Saligrama, and Adam~T
  Kalai. 2016.
\newblock Man is to computer programmer as woman is to homemaker? debiasing
  word embeddings.
\newblock In \emph{Advances in Neural Information Processing Systems}, pages
  4349--4357.

\bibitem[{Borkan et~al.(2019)Borkan, Dixon, Sorensen, Thain, and
  Vasserman}]{borkan2019nuanced}
Daniel Borkan, Lucas Dixon, Jeffrey Sorensen, Nithum Thain, and Lucy Vasserman.
  2019.
\newblock Nuanced metrics for measuring unintended bias with real data for text
  classification.
\newblock In \emph{Companion Proceedings of The 2019 World Wide Web
  Conference}, pages 491--500.

\bibitem[{Breitfeller et~al.(2019)Breitfeller, Ahn, Jurgens, and
  Tsvetkov}]{breitfeller-etal-2019-finding}
Luke Breitfeller, Emily Ahn, David Jurgens, and Yulia Tsvetkov. 2019.
\newblock \href {https://doi.org/10.18653/v1/D19-1176} {Finding
  microaggressions in the wild: A case for locating elusive phenomena in social
  media posts}.
\newblock In \emph{Proceedings of the 2019 Conference on Empirical Methods in
  Natural Language Processing and the 9th International Joint Conference on
  Natural Language Processing (EMNLP-IJCNLP)}, pages 1664--1674, Hong Kong,
  China.

\bibitem[{Caliskan et~al.(2017)Caliskan, Bryson, and
  Narayanan}]{caliskan2017semantics}
Aylin Caliskan, Joanna~J Bryson, and Arvind Narayanan. 2017.
\newblock Semantics derived automatically from language corpora contain
  human-like biases.
\newblock \emph{Science}, 356(6334):183--186.

\bibitem[{Chakrabarty et~al.(2019)Chakrabarty, Gupta, and
  Muresan}]{chakrabarty-etal-2019-pay}
Tuhin Chakrabarty, Kilol Gupta, and Smaranda Muresan. 2019.
\newblock Pay {``}attention{''} to your context when classifying abusive
  language.
\newblock In \emph{Proceedings of the Third Workshop on Abusive Language
  Online}, pages 70--79, Florence, Italy.

\bibitem[{Chandrasekharan et~al.(2018)Chandrasekharan, Samory, Jhaver, Charvat,
  Bruckman, Lampe, Eisenstein, and Gilbert}]{chandrasekharan2018internet}
Eshwar Chandrasekharan, Mattia Samory, Shagun Jhaver, Hunter Charvat, Amy
  Bruckman, Cliff Lampe, Jacob Eisenstein, and Eric Gilbert. 2018.
\newblock The {I}nternet's hidden rules: {A}n empirical study of {R}eddit norm
  violations at micro, meso, and macro scales.
\newblock \emph{Proceedings of the ACM on Human-Computer Interaction},
  2(CSCW):1--25.

\bibitem[{Chatzakou et~al.(2017)Chatzakou, Kourtellis, Blackburn,
  De~Cristofaro, Stringhini, and Vakali}]{chatzakou2017mean}
Despoina Chatzakou, Nicolas Kourtellis, Jeremy Blackburn, Emiliano
  De~Cristofaro, Gianluca Stringhini, and Athena Vakali. 2017.
\newblock Mean birds: Detecting aggression and bullying on {T}witter.
\newblock In \emph{Proceedings of the ACM Conference on Web Science}, pages
  13--22.

\bibitem[{Dadvar et~al.(2013)Dadvar, Trieschnigg, Ordelman, and
  de~Jong}]{dadvar2013improving}
Maral Dadvar, Dolf Trieschnigg, Roeland Ordelman, and Franciska de~Jong. 2013.
\newblock Improving cyberbullying detection with user context.
\newblock In \emph{Proceedings of the European Conference on Information
  Retrieval}, pages 693--696.

\bibitem[{Davidson et~al.(2019)Davidson, Bhattacharya, and
  Weber}]{davidson2019racial}
Thomas Davidson, Debasmita Bhattacharya, and Ingmar Weber. 2019.
\newblock Racial bias in hate speech and abusive language detection datasets.
\newblock In \emph{Proceedings of the Third Workshop on Abusive Language
  Online}, pages 25--35, Florence, Italy.

\bibitem[{Davidson et~al.(2017)Davidson, Warmsley, Macy, and
  Weber}]{davidson2017automated}
Thomas Davidson, Dana Warmsley, Michael Macy, and Ingmar Weber. 2017.
\newblock Automated hate speech detection and the problem of offensive
  language.
\newblock In \emph{Proceedings of the International AAAI Conference on Web and
  Social Media}.

\bibitem[{Dinan et~al.(2020)Dinan, Fan, Wu, Weston, Kiela, and
  Williams}]{dinan2020multi}
Emily Dinan, Angela Fan, Ledell Wu, Jason Weston, Douwe Kiela, and Adina
  Williams. 2020.
\newblock Multi-dimensional gender bias classification.
\newblock \emph{arXiv preprint arXiv:2005.00614}.

\bibitem[{Dixon et~al.(2018)Dixon, Li, Sorensen, Thain, and
  Vasserman}]{dixon2018measuring}
Lucas Dixon, John Li, Jeffrey Sorensen, Nithum Thain, and Lucy Vasserman. 2018.
\newblock Measuring and mitigating unintended bias in text classification.
\newblock In \emph{Proceedings of the 2018 AAAI/ACM Conference on AI, Ethics,
  and Society}, pages 67--73.

\bibitem[{Djuric et~al.(2015)Djuric, Zhou, Morris, Grbovic, Radosavljevic, and
  Bhamidipati}]{djuric2015hate}
Nemanja Djuric, Jing Zhou, Robin Morris, Mihajlo Grbovic, Vladan Radosavljevic,
  and Narayan Bhamidipati. 2015.
\newblock Hate speech detection with comment embeddings.
\newblock In \emph{Proceedings of the International Conference on World Wide
  Web}, pages 29--30.

\bibitem[{Duggan(2017)}]{harassment2017}
Maeve Duggan. 2017.
\newblock \emph{Online Harassment 2017}.
\newblock Pew Research Center.

\bibitem[{Fi{\v{s}}er et~al.(2017)Fi{\v{s}}er, Erjavec, and
  Ljube{\v{s}}i{\'c}}]{fivser2017legal}
Darja Fi{\v{s}}er, Toma{\v{z}} Erjavec, and Nikola Ljube{\v{s}}i{\'c}. 2017.
\newblock Legal framework, dataset and annotation schema for socially
  unacceptable online discourse practices in {S}lovene.
\newblock In \emph{Proceedings of the First Workshop on Abusive Language
  Online}, pages 46--51.

\bibitem[{Flynn and Marley(2014)}]{flynn2014}
T.~N. Flynn and A.~A.~J. Marley. 2014.
\newblock Best-worst scaling: theory and methods.
\newblock In Stephane Hess and Andrew Daly, editors, \emph{Handbook of Choice
  Modelling}, pages 178--201. Edward Elgar Publishing.

\bibitem[{Fortuna and Nunes(2018)}]{fortuna2018survey}
Paula Fortuna and S{\'e}rgio Nunes. 2018.
\newblock A survey on automatic detection of hate speech in text.
\newblock \emph{ACM Computing Surveys (CSUR)}, 51(4):1--30.

\bibitem[{Fortuna et~al.(2020)Fortuna, Soler, and
  Wanner}]{fortuna-etal-2020-toxic}
Paula Fortuna, Juan Soler, and Leo Wanner. 2020.
\newblock \href {https://www.aclweb.org/anthology/2020.lrec-1.838} {Toxic,
  hateful, offensive or abusive? what are we really classifying? an empirical
  analysis of hate speech datasets}.
\newblock In \emph{Proceedings of the 12th Language Resources and Evaluation
  Conference}, pages 6786--6794, Marseille, France.

\bibitem[{Founta et~al.(2019)Founta, Chatzakou, Kourtellis, Blackburn, Vakali,
  and Leontiadis}]{founta2019unified}
Antigoni~Maria Founta, Despoina Chatzakou, Nicolas Kourtellis, Jeremy
  Blackburn, Athena Vakali, and Ilias Leontiadis. 2019.
\newblock A unified deep learning architecture for abuse detection.
\newblock In \emph{Proceedings of the 10th ACM Conference on Web Science},
  pages 105--114.

\bibitem[{Founta et~al.(2018)Founta, Djouvas, Chatzakou, Leontiadis, Blackburn,
  Stringhini, Vakali, Sirivianos, and Kourtellis}]{founta2018large}
Antigoni~Maria Founta, Constantinos Djouvas, Despoina Chatzakou, Ilias
  Leontiadis, Jeremy Blackburn, Gianluca Stringhini, Athena Vakali, Michael
  Sirivianos, and Nicolas Kourtellis. 2018.
\newblock Large scale crowdsourcing and characterization of {T}witter abusive
  behavior.
\newblock In \emph{Proceedings of the International AAAI Conference on Web and
  Social Media}.

\bibitem[{Gao and Huang(2017)}]{gao2017detecting}
Lei Gao and Ruihong Huang. 2017.
\newblock Detecting online hate speech using context aware models.
\newblock In \emph{Proceedings of the International Conference on Recent
  Advances in Natural Language Processing, {RANLP} 2017}, pages 260--266,
  Varna, Bulgaria.

\bibitem[{Garg et~al.(2019)Garg, Perot, Limtiaco, Taly, Chi, and
  Beutel}]{garg2019counterfactual}
Sahaj Garg, Vincent Perot, Nicole Limtiaco, Ankur Taly, Ed~H Chi, and Alex
  Beutel. 2019.
\newblock Counterfactual fairness in text classification through robustness.
\newblock In \emph{Proceedings of the 2019 AAAI/ACM Conference on AI, Ethics,
  and Society}, pages 219--226.

\bibitem[{Gebru et~al.(2018)Gebru, Morgenstern, Vecchione, Vaughan, Wallach,
  Daum{\'e}, and Crawford}]{Gebru2018DatasheetsFD}
Timnit Gebru, J.~Morgenstern, Briana Vecchione, Jennifer~Wortman Vaughan,
  H.~Wallach, Hal Daum{\'e}, and K.~Crawford. 2018.
\newblock Datasheets for datasets.
\newblock In \emph{Proceedings of the 5th Workshop on Fairness, Accountability,
  and Transparency in Machine Learning}, Stockholm, Sweden.

\bibitem[{Gehman et~al.(2020)Gehman, Gururangan, Sap, Choi, and
  Smith}]{gehman2020realtoxicityprompts}
Sam Gehman, Suchin Gururangan, Maarten Sap, Yejin Choi, and Noah~A Smith. 2020.
\newblock Realtoxicityprompts: Evaluating neural toxic degeneration in language
  models.
\newblock In \emph{Findings of EMNLP}.

\bibitem[{Gitari et~al.(2015)Gitari, Zuping, Damien, and
  Long}]{gitari2015lexicon}
Njagi~Dennis Gitari, Zhang Zuping, Hanyurwimfura Damien, and Jun Long. 2015.
\newblock A lexicon-based approach for hate speech detection.
\newblock \emph{International Journal of Multimedia and Ubiquitous
  Engineering}, 10(4):215--230.

\bibitem[{Goffin and Olson(2011)}]{goffin2011all}
Richard~D. Goffin and James~M. Olson. 2011.
\newblock Is it all relative? {C}omparative judgments and the possible
  improvement of self-ratings and ratings of others.
\newblock \emph{Perspectives on Psychological Science}, 6(1):48--60.

\bibitem[{Golbeck et~al.(2017)Golbeck, Ashktorab, Banjo, Berlinger, Bhagwan,
  Buntain, Cheakalos, Geller, Gnanasekaran, Gunasekaran
  et~al.}]{golbeck2017large}
Jennifer Golbeck, Zahra Ashktorab, Rashad~O Banjo, Alexandra Berlinger,
  Siddharth Bhagwan, Cody Buntain, Paul Cheakalos, Alicia~A Geller,
  Rajesh~Kumar Gnanasekaran, Raja~Rajan Gunasekaran, et~al. 2017.
\newblock A large labeled corpus for online harassment research.
\newblock In \emph{Proceedings of the ACM Conference on Web Science}, pages
  229--233.

\bibitem[{Gr{\"o}ndahl et~al.(2018)Gr{\"o}ndahl, Pajola, Juuti, Conti, and
  Asokan}]{grondahl2018all}
Tommi Gr{\"o}ndahl, Luca Pajola, Mika Juuti, Mauro Conti, and N~Asokan. 2018.
\newblock All you need is ``love'' evading hate speech detection.
\newblock In \emph{Proceedings of the 11th ACM Workshop on Artificial
  Intelligence and Security}, pages 2--12.

\bibitem[{Gunasekara and Nejadgholi(2018)}]{gunasekara2018review}
Isuru Gunasekara and Isar Nejadgholi. 2018.
\newblock A review of standard text classification practices for multi-label
  toxicity identification of online content.
\newblock In \emph{Proceedings of the 2nd Workshop on Abusive Language Online
  (ALW2)}, pages 21--25.

\bibitem[{Hinduja and Patchin(2020)}]{cyberbullying2020}
S.~Hinduja and J.~W. Patchin. 2020.
\newblock \emph{Cyberbullying fact sheet: Identification, Prevention, and
  Response}.
\newblock Cyberbullying Research Center.

\bibitem[{Hosseini et~al.(2017)Hosseini, Kannan, Zhang, and
  Poovendran}]{hosseini2017deceiving}
Hossein Hosseini, Sreeram Kannan, Baosen Zhang, and Radha Poovendran. 2017.
\newblock Deceiving {G}oogle's {P}erspective {API} built for detecting toxic
  comments.
\newblock \emph{arXiv preprint arXiv:1702.08138}.

\bibitem[{Hosseinmardi et~al.(2015)Hosseinmardi, Mattson, Rafiq, Han, Lv, and
  Mishra}]{hosseinmardi2015analyzing}
Homa Hosseinmardi, Sabrina~Arredondo Mattson, Rahat~Ibn Rafiq, Richard Han, Qin
  Lv, and Shivakant Mishra. 2015.
\newblock Analyzing labeled cyberbullying incidents on the {I}nstagram social
  network.
\newblock In \emph{Proceedings of the International Conference on Social
  Informatics}, pages 49--66.

\bibitem[{Jain and Wallace(2019)}]{jain-wallace-2019-attention}
Sarthak Jain and Byron~C. Wallace. 2019.
\newblock {A}ttention is not {E}xplanation.
\newblock In \emph{Proceedings of the 2019 Conference of the North {A}merican
  Chapter of the Association for Computational Linguistics: Human Language
  Technologies, Volume 1 (Long and Short Papers)}, pages 3543--3556,
  Minneapolis, Minnesota.

\bibitem[{{Jigsaw, Perspective {API}}(2017)}]{PersAPI}
{Jigsaw, Perspective {API}}. 2017.
\newblock \url{https://www.perspectiveapi.com/#/home}.

\bibitem[{Jurgens et~al.(2019)Jurgens, Hemphill, and
  Chandrasekharan}]{jurgens-etal-2019-just}
David Jurgens, Libby Hemphill, and Eshwar Chandrasekharan. 2019.
\newblock A just and comprehensive strategy for using {NLP} to address online
  abuse.
\newblock In \emph{Proceedings of the 57th Annual Meeting of the Association
  for Computational Linguistics}, pages 3658--3666, Florence, Italy.

\bibitem[{Kalin et~al.(2020)Kalin, Noever, and Dozier}]{kalin2020systematic}
Josh Kalin, David Noever, and Gerry Dozier. 2020.
\newblock Systematic attack surface reduction for deployed sentiment analysis
  models.
\newblock \emph{arXiv preprint arXiv:2006.11130}.

\bibitem[{Kiritchenko and Mohammad(2016)}]{maxdiff-naacl2016}
Svetlana Kiritchenko and Saif~M. Mohammad. 2016.
\newblock Capturing reliable fine-grained sentiment associations by
  crowdsourcing and best--worst scaling.
\newblock In \emph{Proceedings of the Annual Conference of the North American
  Chapter of the Association for Computational Linguistics: Human Language
  Technologies (NAACL)}, San Diego, California.

\bibitem[{Kiritchenko and Mohammad(2017)}]{KiritchenkoM2017bwsvsrs}
Svetlana Kiritchenko and Saif~M. Mohammad. 2017.
\newblock Best-worst scaling more reliable than rating scales: A case study on
  sentiment intensity annotation.
\newblock In \emph{Proceedings of the Annual Meeting of the Association for
  Computational Linguistics (ACL)}, Vancouver, Canada.

\bibitem[{Kumar et~al.(2018)Kumar, Ojha, Malmasi, and
  Zampieri}]{kumar2018benchmarking}
Ritesh Kumar, Atul~Kr Ojha, Shervin Malmasi, and Marcos Zampieri. 2018.
\newblock Benchmarking aggression identification in social media.
\newblock In \emph{Proceedings of the First Workshop on Trolling, Aggression
  and Cyberbullying (TRAC-2018)}, pages 1--11.

\bibitem[{Kurita et~al.(2020)Kurita, Michel, and Neubig}]{kurita2020weight}
Keita Kurita, Paul Michel, and Graham Neubig. 2020.
\newblock Weight poisoning attacks on pre-trained models.
\newblock In \emph{Proceedings of the Annual Meeting of the Association for
  Computational Linguistics}.

\bibitem[{Kurita et~al.(2019)Kurita, Vyas, Pareek, Black, and
  Tsvetkov}]{kurita2019measuring}
Keita Kurita, Nidhi Vyas, Ayush Pareek, Alan~W Black, and Yulia Tsvetkov. 2019.
\newblock Measuring bias in contextualized word representations.
\newblock In \emph{Proceedings of the First Workshop on Gender Bias in Natural
  Language Processing}, pages 166--172.

\bibitem[{Louviere et~al.(2015)Louviere, Flynn, and Marley}]{Louviere2015}
Jordan~J. Louviere, Terry~N. Flynn, and A.~A.~J. Marley. 2015.
\newblock \emph{{Best-Worst Scaling}: Theory, Methods and Applications}.
\newblock Cambridge University Press.

\bibitem[{Louviere and Woodworth(1990)}]{Louviere_1990}
Jordan~J. Louviere and George~G. Woodworth. 1990.
\newblock Best-worst analysis.
\newblock Working Paper.
\newblock Department of Marketing and Economic Analysis, University of Alberta.

\bibitem[{Mahajan et~al.(2020)Mahajan, Shah, and
  Jafar}]{mahajan2020explainable}
Aditya Mahajan, Divyank Shah, and Gibraan Jafar. 2020.
\newblock Explainable {AI} approach towards toxic comment classification.
\newblock Technical report.

\bibitem[{Manzini et~al.(2019)Manzini, Chong, Black, and
  Tsvetkov}]{manzini2019black}
Thomas Manzini, Lim~Yao Chong, Alan~W Black, and Yulia Tsvetkov. 2019.
\newblock Black is to criminal as caucasian is to police: Detecting and
  removing multiclass bias in word embeddings.
\newblock In \emph{Proceedings of the 2019 Conference of the North American
  Chapter of the Association for Computational Linguistics: Human Language
  Technologies, Volume 1 (Long and Short Papers)}, pages 615--621.

\bibitem[{May et~al.(2019)May, Wang, Bordia, Bowman, and
  Rudinger}]{may2019measuring}
Chandler May, Alex Wang, Shikha Bordia, Samuel Bowman, and Rachel Rudinger.
  2019.
\newblock On measuring social biases in sentence encoders.
\newblock In \emph{Proceedings of the 2019 Conference of the North American
  Chapter of the Association for Computational Linguistics: Human Language
  Technologies, Volume 1 (Long and Short Papers)}, pages 622--628.

\bibitem[{Mishra et~al.(2019)Mishra, Yannakoudakis, and
  Shutova}]{mishra2019tackling}
Pushkar Mishra, Helen Yannakoudakis, and Ekaterina Shutova. 2019.
\newblock Tackling online abuse: A survey of automated abuse detection methods.
\newblock \emph{arXiv preprint arXiv:1908.06024}.

\bibitem[{Mitchell et~al.(2019)Mitchell, Wu, Zaldivar, Barnes, Vasserman,
  Hutchinson, Spitzer, Raji, and Gebru}]{mitchell2019model}
Margaret Mitchell, Simone Wu, Andrew Zaldivar, Parker Barnes, Lucy Vasserman,
  Ben Hutchinson, Elena Spitzer, Inioluwa~Deborah Raji, and Timnit Gebru. 2019.
\newblock Model cards for model reporting.
\newblock In \emph{Proceedings of the Conference on Fairness, Accountability,
  and Transparency}, pages 220--229.

\bibitem[{Nadeem et~al.(2020)Nadeem, Bethke, and Reddy}]{nadeem2020stereoset}
Moin Nadeem, Anna Bethke, and Siva Reddy. 2020.
\newblock Stereoset: Measuring stereotypical bias in pretrained language
  models.
\newblock \emph{arXiv preprint arXiv:2004.09456}.

\bibitem[{Nangia et~al.(2020)Nangia, Vania, Bhalerao, and
  Bowman}]{nangia2020crows}
Nikita Nangia, Clara Vania, Rasika Bhalerao, and Samuel~R. Bowman. 2020.
\newblock Crows-pairs: A challenge dataset for measuring social biases in
  masked language models.
\newblock In \emph{Proceedings of the Conference on Empirical Methods in
  Natural Language Processing (EMNLP)}.

\bibitem[{Nejadgholi and Kiritchenko(2020)}]{nejadgholi2020cross}
Isar Nejadgholi and Svetlana Kiritchenko. 2020.
\newblock On cross-dataset generalization in automatic detection of online
  abuse.
\newblock In \emph{Proceedings of the 4th Workshop on Online Abuse and Harms}.

\bibitem[{Niemann et~al.(2019)Niemann, Riehle, Brunk, and
  Becker}]{niemann2019abusive}
Marco Niemann, Dennis~M Riehle, Jens Brunk, and J{\"o}rg Becker. 2019.
\newblock What is abusive language?
\newblock In \emph{Proceedings of the Multidisciplinary International Symposium
  on Disinformation in Open Online Media}, pages 59--73.

\bibitem[{Nobata et~al.(2016)Nobata, Tetreault, Thomas, Mehdad, and
  Chang}]{nobata2016abusive}
Chikashi Nobata, Joel Tetreault, Achint Thomas, Yashar Mehdad, and Yi~Chang.
  2016.
\newblock Abusive language detection in online user content.
\newblock In \emph{Proceedings of the International Conference on World Wide
  Web}, pages 145--153.

\bibitem[{Park et~al.(2018)Park, Shin, and Fung}]{park2018reducing}
Ji~Ho Park, Jamin Shin, and Pascale Fung. 2018.
\newblock Reducing gender bias in abusive language detection.
\newblock In \emph{Proceedings of the Conference on Empirical Methods in
  Natural Language Processing}, pages 2799--2804, Brussels, Belgium.

\bibitem[{Pohjonen and Udupa(2017)}]{pohjonen2017extreme}
Matti Pohjonen and Sahana Udupa. 2017.
\newblock Extreme speech online: An anthropological critique of hate speech
  debates.
\newblock \emph{International Journal of Communication}, 11:19.

\bibitem[{Poletto et~al.(2017)Poletto, Stranisci, Sanguinetti, Patti, and
  Bosco}]{poletto2017hate}
Fabio Poletto, Marco Stranisci, Manuela Sanguinetti, Viviana Patti, and
  Cristina Bosco. 2017.
\newblock Hate speech annotation: {A}nalysis of an {I}talian {T}witter corpus.
\newblock In \emph{Proceedings of the Italian Conference on Computational
  Linguistics (CLiC-it 2017)}, pages 1--6.

\bibitem[{Presser and Schuman(1996)}]{presser2004questions}
Stanley Presser and Howard Schuman. 1996.
\newblock \emph{Questions and Answers in Attitude Surveys: Experiments on
  Question Form, Wording, and Context}.
\newblock SAGE Publications, Inc.

\bibitem[{Razavi et~al.(2010)Razavi, Inkpen, Uritsky, and
  Matwin}]{razavi2010offensive}
Amir~H Razavi, Diana Inkpen, Sasha Uritsky, and Stan Matwin. 2010.
\newblock Offensive language detection using multi-level classification.
\newblock In \emph{Proceedings of the Canadian Conference on Artificial
  Intelligence}, pages 16--27.

\bibitem[{Ribeiro et~al.(2016)Ribeiro, Singh, and Guestrin}]{ribeiro2016should}
Marco~Tulio Ribeiro, Sameer Singh, and Carlos Guestrin. 2016.
\newblock ``{W}hy should {I} trust you?'' {E}xplaining the predictions of any
  classifier.
\newblock In \emph{Proceedings of the ACM SIGKDD International Conference on
  Knowledge Discovery and Data Mining}, pages 1135--1144.

\bibitem[{Roberts(2016)}]{roberts2016commercial}
Sarah~T. Roberts. 2016.
\newblock Commercial content moderation: Digital laborers' dirty work.
\newblock In \emph{The Intersectional Internet: Race, Sex, Class and Culture
  Online}. Peter Lang Publishing.

\bibitem[{{Salawu} et~al.(2020){Salawu}, {He}, and {Lumsden}}]{Salawu2020}
S.~{Salawu}, Y.~{He}, and J.~{Lumsden}. 2020.
\newblock Approaches to automated detection of cyberbullying: A survey.
\newblock \emph{IEEE Transactions on Affective Computing}, 11(1):3--24.

\bibitem[{Sap et~al.(2019)Sap, Card, Gabriel, Choi, and Smith}]{sap2019risk}
Maarten Sap, Dallas Card, Saadia Gabriel, Yejin Choi, and Noah~A Smith. 2019.
\newblock The risk of racial bias in hate speech detection.
\newblock In \emph{Proceedings of the 57th Annual Meeting of the Association
  for Computational Linguistics}, pages 1668--1678.

\bibitem[{Sap et~al.(2020)Sap, Gabriel, Qin, Jurafsky, Smith, and
  Choi}]{sap2020socialbiasframes}
Maarten Sap, Saadia Gabriel, Lianhui Qin, Dan Jurafsky, Noah~A Smith, and Yejin
  Choi. 2020.
\newblock Social bias frames: Reasoning about social and power implications of
  language.
\newblock In \emph{Proceedings of the Annual Meeting of the Association for
  Computational Linguistics}.

\bibitem[{Schmidt and Wiegand(2017)}]{schmidt2017survey}
Anna Schmidt and Michael Wiegand. 2017.
\newblock A survey on hate speech detection using natural language processing.
\newblock In \emph{Proceedings of the Fifth International Workshop on Natural
  Language Processing for Social Media}, pages 1--10.

\bibitem[{Shepherd et~al.(2015)Shepherd, Harvey, Jordan, Srauy, and
  Miltner}]{shepherd2015histories}
Tamara Shepherd, Alison Harvey, Tim Jordan, Sam Srauy, and Kate Miltner. 2015.
\newblock Histories of hating.
\newblock \emph{Social Media+ Society}, 1(2).

\bibitem[{Spertus(1997)}]{spertus1997smokey}
Ellen Spertus. 1997.
\newblock Smokey: Automatic recognition of hostile messages.
\newblock In \emph{Proceedings of the AAAI Conference on Innovative
  Applications of Artificial Intelligence}, pages 1058--1065.

\bibitem[{Srivastava and Khurana(2019)}]{srivastava2019detecting}
Saurabh Srivastava and Prerna Khurana. 2019.
\newblock Detecting aggression and toxicity using a multi dimension capsule
  network.
\newblock In \emph{Proceedings of the Third Workshop on Abusive Language
  Online}, pages 157--162.

\bibitem[{Tversky and Kahneman(1974)}]{tversky1974judgment}
Amos Tversky and Daniel Kahneman. 1974.
\newblock Judgment under uncertainty: Heuristics and biases.
\newblock \emph{Science}, 185(4157):1124--1131.

\bibitem[{Van~Bruwaene et~al.(2020)Van~Bruwaene, Huang, and
  Inkpen}]{van2020multi}
David Van~Bruwaene, Qianjia Huang, and Diana Inkpen. 2020.
\newblock A multi-platform dataset for detecting cyberbullying in social media.
\newblock \emph{Language Resources and Evaluation}, pages 1--24.

\bibitem[{Vidgen and Derczynski(2020)}]{vidgen2020directions}
Bertie Vidgen and Leon Derczynski. 2020.
\newblock Directions in abusive language training data: Garbage in, garbage
  out.
\newblock \emph{arXiv preprint arXiv:2004.01670}.

\bibitem[{Vidgen et~al.(2019)Vidgen, Harris, Nguyen, Tromble, Hale, and
  Margetts}]{vidgen-etal-2019-challenges}
Bertie Vidgen, Alex Harris, Dong Nguyen, Rebekah Tromble, Scott Hale, and Helen
  Margetts. 2019.
\newblock Challenges and frontiers in abusive content detection.
\newblock In \emph{Proceedings of the Third Workshop on Abusive Language
  Online}, pages 80--93, Florence, Italy.

\bibitem[{Waseem(2016)}]{waseem2016you}
Zeerak Waseem. 2016.
\newblock Are you a racist or am {I} seeing things? {A}nnotator influence on
  hate speech detection on {T}witter.
\newblock In \emph{Proceedings of the First Workshop on NLP and Computational
  Social Science}, pages 138--142.

\bibitem[{Waseem et~al.(2017)Waseem, Davidson, Warmsley, and
  Weber}]{waseem-etal-2017-understanding}
Zeerak Waseem, Thomas Davidson, Dana Warmsley, and Ingmar Weber. 2017.
\newblock Understanding abuse: A typology of abusive language detection
  subtasks.
\newblock In \emph{Proceedings of the First Workshop on Abusive Language
  Online}, pages 78--84, Vancouver, BC, Canada.

\bibitem[{Waseem and Hovy(2016)}]{waseem2016hateful}
Zeerak Waseem and Dirk Hovy. 2016.
\newblock Hateful symbols or hateful people? {P}redictive features for hate
  speech detection on {T}witter.
\newblock In \emph{Proceedings of the NAACL Student Research Workshop}, pages
  88--93.

\bibitem[{Waseem et~al.(2018)Waseem, Thorne, and Bingel}]{waseem2018bridging}
Zeerak Waseem, James Thorne, and Joachim Bingel. 2018.
\newblock Bridging the gaps: Multi task learning for domain transfer of hate
  speech detection.
\newblock In Golbeck J., editor, \emph{Online harassment}, pages 29--55.
  Springer.

\bibitem[{Wiegand et~al.(2019)Wiegand, Ruppenhofer, and
  Kleinbauer}]{wiegand2019detection}
Michael Wiegand, Josef Ruppenhofer, and Thomas Kleinbauer. 2019.
\newblock Detection of abusive language: the problem of biased datasets.
\newblock In \emph{Proceedings of the Conference of the North American Chapter
  of the Association for Computational Linguistics: Human Language
  Technologies}, pages 602--608.

\bibitem[{Wiegand et~al.(2018)Wiegand, Ruppenhofer, Schmidt, and
  Greenberg}]{wiegand2018inducing}
Michael Wiegand, Josef Ruppenhofer, Anna Schmidt, and Clayton Greenberg. 2018.
\newblock Inducing a lexicon of abusive words--a feature-based approach.
\newblock In \emph{Proceedings of the 2018 Conference of the North American
  Chapter of the Association for Computational Linguistics: Human Language
  Technologies}, pages 1046--1056.

\bibitem[{Wiegreffe and Pinter(2019)}]{wiegreffe2019attention}
Sarah Wiegreffe and Yuval Pinter. 2019.
\newblock Attention is not not explanation.
\newblock In \emph{Proceedings of the 2019 Conference on Empirical Methods in
  Natural Language Processing and the 9th International Joint Conference on
  Natural Language Processing (EMNLP-IJCNLP)}, pages 11--20.

\bibitem[{Wilhelm and Joeckel(2019)}]{wilhelm2019gendered}
Claudia Wilhelm and Sven Joeckel. 2019.
\newblock Gendered morality and backlash effects in online discussions: An
  experimental study on how users respond to hate speech comments against women
  and sexual minorities.
\newblock \emph{Sex Roles}, 80(7-8):381--392.

\bibitem[{Wulczyn et~al.(2017)Wulczyn, Thain, and Dixon}]{wulczyn2017ex}
Ellery Wulczyn, Nithum Thain, and Lucas Dixon. 2017.
\newblock Ex machina: Personal attacks seen at scale.
\newblock In \emph{Proceedings of the 26th International Conference on World
  Wide Web}, pages 1391--1399.

\bibitem[{Zampieri et~al.(2019{\natexlab{a}})Zampieri, Malmasi, Nakov,
  Rosenthal, Farra, and Kumar}]{zampieri-etal-2019-predicting}
Marcos Zampieri, Shervin Malmasi, Preslav Nakov, Sara Rosenthal, Noura Farra,
  and Ritesh Kumar. 2019{\natexlab{a}}.
\newblock Predicting the type and target of offensive posts in social media.
\newblock In \emph{Proceedings of the Conference of the North {A}merican
  Chapter of the Association for Computational Linguistics: Human Language
  Technologies}, pages 1415--1420, Minneapolis, Minnesota.

\bibitem[{Zampieri et~al.(2019{\natexlab{b}})Zampieri, Malmasi, Nakov,
  Rosenthal, Farra, and Kumar}]{zampieri2019semeval}
Marcos Zampieri, Shervin Malmasi, Preslav Nakov, Sara Rosenthal, Noura Farra,
  and Ritesh Kumar. 2019{\natexlab{b}}.
\newblock Semeval-2019 task 6: Identifying and categorizing offensive language
  in social media (offenseval).
\newblock In \emph{Proceedings of the 13th International Workshop on Semantic
  Evaluation}, pages 75--86.

\bibitem[{Zufall et~al.(2020)Zufall, Zhang, Kloppenborg, and
  Zesch}]{zufall2020operationalizing}
Frederike Zufall, Huangpan Zhang, Katharina Kloppenborg, and Torsten Zesch.
  2020.
\newblock Operationalizing the legal concept of `incitement to hatred' as an
  {NLP} task.
\newblock \emph{arXiv preprint arXiv:2004.03422}.

\end{thebibliography}

\end{document}